\newcommand{\tridentata}{\textit{L. tridentata}}

\documentclass[aps,prl,twocolumn,groupedaddress]{revtex4-2}

\usepackage{graphicx}
\usepackage{amsmath}

\usepackage{makecell}

\usepackage{xcolor}
\usepackage{soul}
\usepackage{bm}
\usepackage{svg}
\usepackage{float}

\begin{document}

\title{Pacific Lamprey Inspired Climbing}

\author{Brian~Van Stratum} 
\author{Kourosh~Shoele}
\author{Jonathan E.~Clark}
\affiliation{Department of Mechanical Engineering, FAMU-FSU College of Engineering, Tallahassee, FL, 32310, USA}

\date{\today}

\begin{abstract}
Snakes and their bio-inspired robot counterparts have demonstrated locomotion on a wide range of terrains. However, dynamic vertical climbing is one locomotion strategy that has received little attention in the existing snake robotics literature. We demonstrate a new scansorial gait and robot inspired by the locomotion of the Pacific Lamprey. This new gait allows a robot to steer while climbing on flat, near-vertical surfaces. A reduced-order model is developed and used to explore the relationship between body actuation and vertical and lateral motions of the robot. Trident, the new wall climbing lamprey-inspired robot, demonstrates dynamic climbing on flat vertical surfaces with a peak net vertical stride displacement of 4.1 cm per step. Actuating at 1.3 Hz, Trident attains a vertical climbing speed of 4.8 cm/s (0.09 Bl/s) at specific resistance of 8.3. Trident can also traverse laterally at 9 cm/s (0.17 Bl/s). Moreover, Trident is able to make 14\% longer strides than the Pacific Lamprey when climbing vertically. The computational and experimental results demonstrate that a lamprey-inspired climbing gait coupled with appropriate attachment is a useful climbing strategy for snake robots climbing near vertical surfaces with limited push points.
\end{abstract}

\maketitle

\section{INTRODUCTION}
 {Snakes}, limbless lizards and other long slender organisms are incredibly adaptable to various terrains. Incorporating diverse locomotion gaits or modes, they are able to move on or through media as diverse as sand, water, trees, and even aerial environments \cite{gray1946mechanism,maladen2009undulatory,astley2007effects,socha2002gliding}. Starting with Hirose's Active Cord Mechanisms, roboticists have utilized inspiration from snakes in their designs for almost 50 years\cite{hirose2004biologically}. The snake's mobility in various terrain and their long slender body morphology have interested researchers in designing bio-inspired robots for challenging applications in search and rescue, surveillance, and exploration operations \cite{kamegawa2004development,whitman2018snake}. These snake-like robots can navigate a diverse range of environments such as sandy slopes \cite{marvi2014sidewinding} was well as in the water \cite{mcisaac2003motion} and vegetated areas \cite{tesch2009parameterized}. Like their biological counterparts, they can push against obstacles in their environment for locomotion\cite{gasc1989propulsive,transeth2008snake}. Snake robots also exhibit amphibious multi-modal functionality \cite{crespi2006amphibot,hirose2009snake} as well as the ability to climb cracks, ladders, and poles. Still, they are not yet designed for climbing shear walls with performance comparable to their legged counterparts.  \cite{shapiro2007frictional,takemori2018ladder,lipkin2007differentiable,tesch2009parameterized}.

   \begin{figure}[t]
      \centering
      \includegraphics[width = \columnwidth]{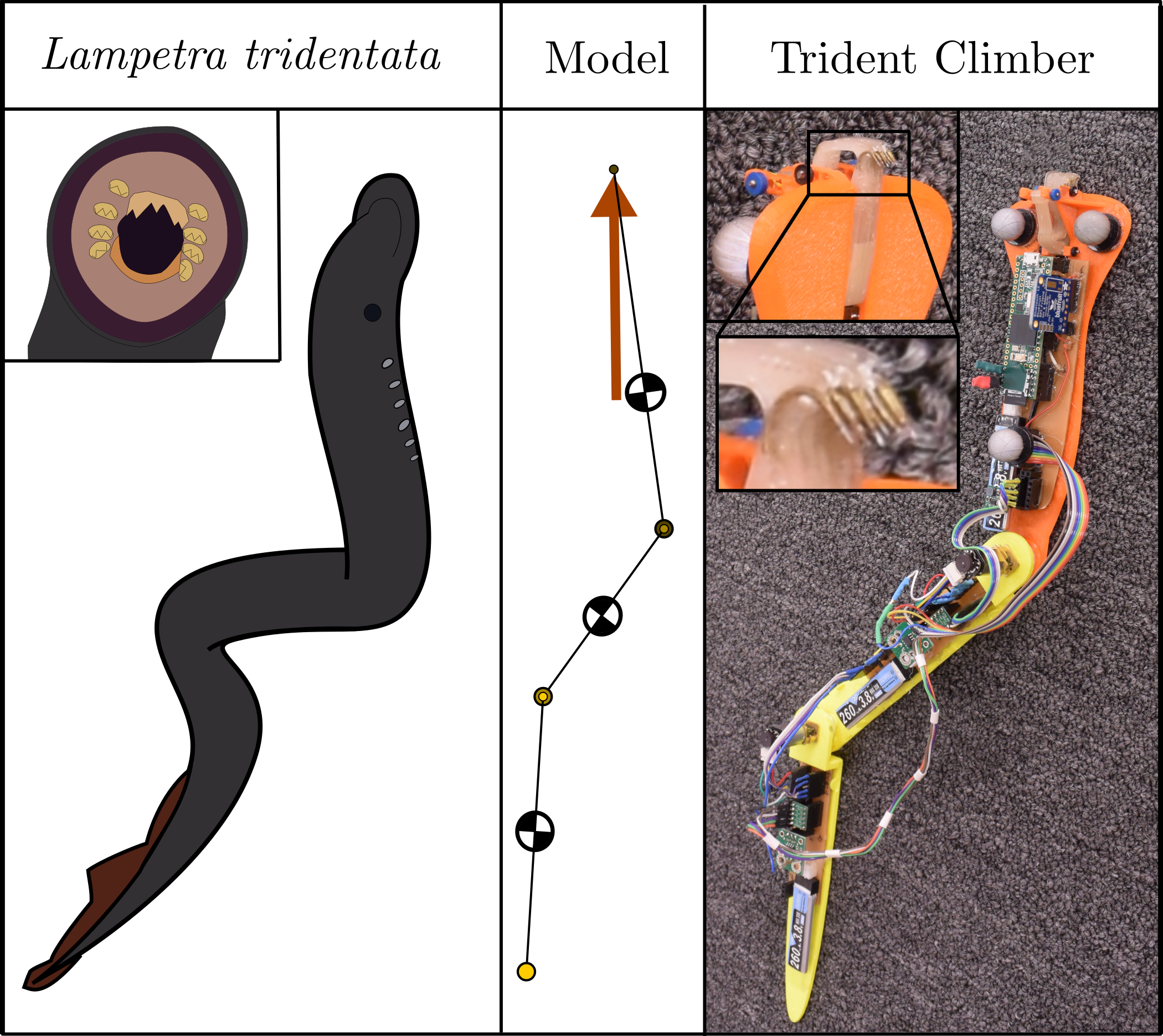}
      \caption{\textit{ L. tridentata} or the Pacific Lamprey climbs vertical surfaces by compressing and extending its body. The main characteristics of their dynamic motion are represented using a reduced-order model. Which is employed to design and steer Trident Climber. \textit{L. tridentata} have an oral disk and buccal funnel for generating adhesion Trident Climber's analogous feature is a microspine array.}
      \label{concept}
   \end{figure}

Legged wall-climbing robots are oftentimes designed with the help of reduced-order dynamic models which enables  such robots to reach increased speed, agility, efficiency, and become robust to disturbances \cite{dickson2012effect,lynch2012bioinspired,provancher2010rocr}. On the other hand, much less focus is put on dynamic locomotion in terrestrial snake robots. Snakes in general, and robots moving in a snake-like way, experience low inertial forces during terrestrial locomotion \cite{hu2009mechanics}. As a result, their performance is typically dependent only on the instantaneous state of the robot, and kinematic or geometric control approaches are effective. Notable exceptions to this generalization occur when studying  locomotion in the air of the gliding snake \textit{Chrsopelea paradisi} \cite{socha20053,jafari2017control} and numerical exploration of sinus lifting, sidewinding, and lateral undulation modes for snake robots \cite{ariizumi2017dynamic}.

Similarly, it is common to make a quasi-static assumption to model snake robots while climbing. Inchworm-inspired climbing locomotion utilizing suction-based adhesion has shown to be effective for traversing cylindrical and glass domains \cite{guan2012modular}. Another approach for climbing cylindrical obstacles is to utilize the many degrees of freedom of snake robots to bend around a pole forming the snake body into a helix while rolling on its longitudinal axis \cite{tesch2009parameterized,hatton2010generating,rollinson2013gait}. Shift control is another snake robot control method used for climbing which employs predetermined bending shapes designed for structured environmental features such as ladder rungs or regularly spaced holds. These contrived shapes propagate posteriorly along a climbing snake-like robot's body\cite{hirose2004biologically,takemori2018ladder}. On the other hand, nature provides an example of dynamic and agile snake-like climbing in the Pacific Lamprey.

 The Pacific Lamprey or \textit{ Lampetra tridentata}, is a fish with a unique capacity for climbing. {tridentata} holds on to vertical surfaces with its oral disk, and moves its body, propagating body curvature from anterior to posterior in an anguilliform swimming motion. Kemp, Tsuzaki and Mosser documented the kinematics of the climbing of the Pacific Lamprey \cite{kemp2009linking}. They report that Lamprey climbing involves, ``intermittent bouts of activity where compression of the body alternates with rapid straitening resulting in vertical progress". This type of body actuation allows the lamprey to progressively climb up vertical barriers such as dams, weirs, and waterfalls over wet, sharp, and smooth surfaces. Moreover, lampreys use this same body actuation to move laterally while climbing. A Numerical simulation by Zhu and Kemp later investigated the climbing of the Pacific Lamprey as a continuous flexible beam\cite{zhu2011numerical}. They discovered that kinematic body undulation reported by Kemp is associated with optimal climbing efficiency.  

We note that current snake robots are incapable of climbing steep, smooth surfaces in a lamprey-like way. We hypothesize that a feedforward contraction and expansion control policy can produce bio-inspired climbing locomotion in a simple multi-link segmented robot. We explore this hypothesis with a new robot designed using a reduced-order dynamic model. A special attachment mechanism is designed to mimic the unique purchase-keeping function of \textit{L. tridentata's} head using a microspine array. Figure \ref{concept} illustrates our design process moving from the biological inspiration of \textit{L. tridentata} to the robot Trident Climber using the reduced-order serial linkage model. 
 
In section \ref{sec:ReducedOrderModeling} of this paper, we present our hybrid dynamic mathematical and numerical model of Lamprey and show how insights from the model were used to design the Trident robot. We explain our experimental method  In section \ref{ExperimentalSection}. In Section \ref{results} we present results from the experimental and numerical studies. In section \ref{sec:LocomotionStrategies} we apply these results and demonstrate steering a lamprey-like climbing robot. We also demonstrate that Trident performs favorably when compared with the animal exemplar. Section \ref{sec:conclusions} finishes with some conclusions and considers future work.

\begin{figure}[t]
      \centering
      \includegraphics[width = \columnwidth]{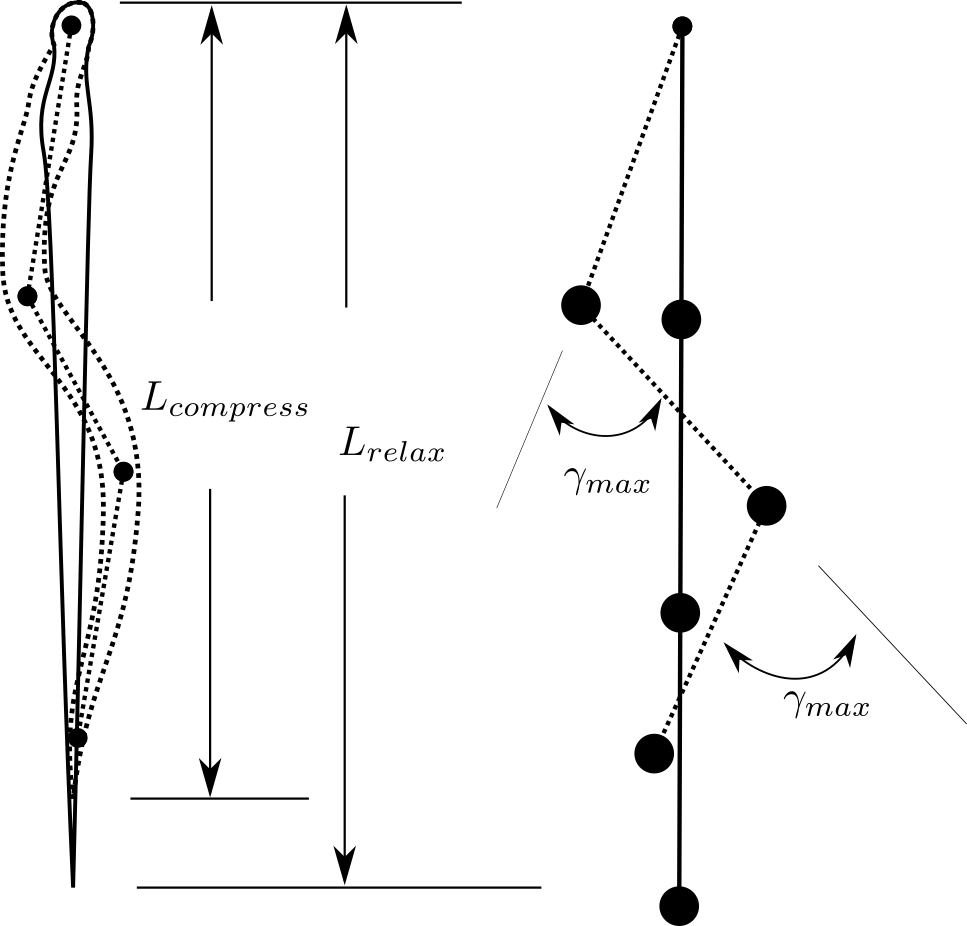}
      \caption{When lampreys are climbing a mean body compression ratio of 12\% has been measured. At this compression, the body of the lamprey shows between one and two curvature waves (one wave is shown in the figure.) We propose to minimally approximate a single body curvature wave with a three-piece serial linkage. Joint angle $\gamma$ is a corresponding measure of compression for a three-piece serial linkage (when both joint angles are held $180^o$ out of phase.)}
      \label{gammaDef}
   \end{figure}
\section{Reduced Order Modeling}
\label{sec:ReducedOrderModeling}
 A computationally tractable model that captures salient aspects of the dynamics allows insights from physics to be encoded into a robot design and provides knowledge of animal locomotion.  It has been shown that the climbing of cockroaches and geckos can be modeled with a simplified climbing model or template \cite{goldman2006dynamics}. In the spirit of these models, we propose to model the continuum body of the lamprey using serial planar linkages. As shown in fig. \ref{gammaDef}, three links are chosen because it is the minimum needed to reproduce the curvature wave observed in climbing lamprey. Reducing this continuum system to a small number of degrees of freedom allows a detailed study of the robot's performance with a limited number of actuators.

\label{ROmodel}
   \begin{figure}[t]
      \centering
      \includegraphics[width = \columnwidth]{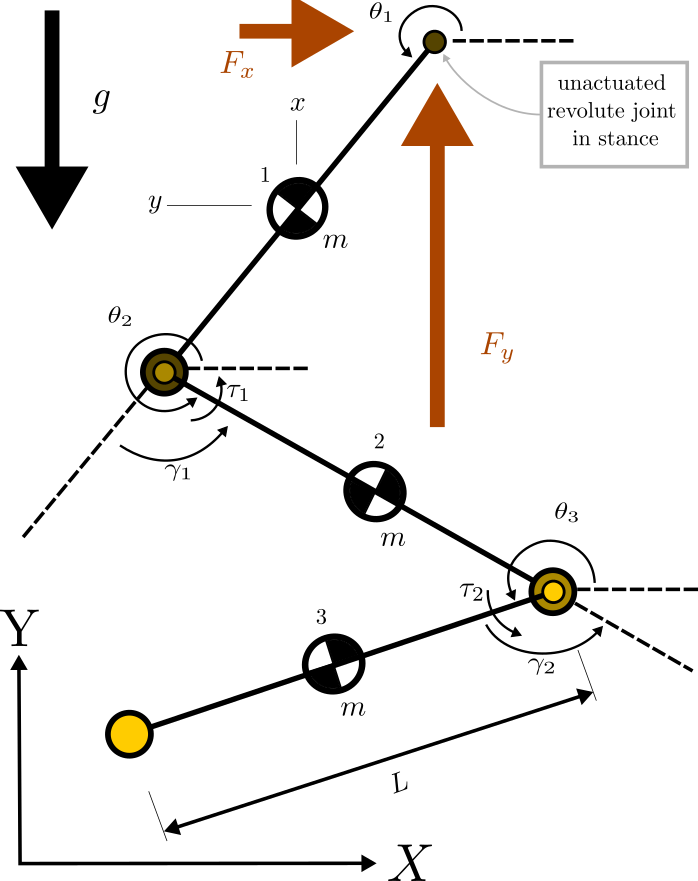}
      \caption{Mechanism diagram depicting the analytical model for \textit{L. tridentata} inspired climbing. Conventional definitions for inertial link angles $\theta$ and relative joint angle $\gamma$ are shown. Reaction forces ($F_x$ and $F_y$) at the head are not present during the flight phase.}
      \label{jumperFBD}
   \end{figure}
Figure \ref{jumperFBD} shows a schematic of the computational model used in our study. The lamprey head is modeled as a revolute joint with no actuation torque. Body muscle actuation is represented by torques applied at the second and third revolute joints. The parameters of this model are tabulated in section \ref{sec:RobotPlatform}. 
We model lamprey climbing in two separate phases: an attached phase (stance phase) when its oral disk is applying suction force on the surface and a flight phase when it is moving only under the influence of gravity. In order to model these two phases, a hybrid dynamic model is proposed.

\subsection{Stance}
The Newton-Euler method is used to derive the dynamic equations of motion and to calculate the internal forces required for robot design and dynamic equation transition. Equations for the $ith$ link in series are given by Equation \ref{eq:NewtonEuler}.

\begin{equation}
\label{eq:NewtonEuler}
\begin{aligned}
Fx_i-Fx_{i+1}-\text{sgn}(\dot{x}_i)\,C_x &=m\,\ddot{x_i}\\
-m\,g\,\cos(\alpha)+Fy_i-Fy_{i+1}-\text{sgn}(\dot{y}_i)\,C_y &=m\,\ddot{y_i}\\
\frac{L}{2}\left(Fy_i+Fy_{i+1}-Fx_i-Fx_{i+1}\right)+\tau_{i-1}-\tau_{i} &= \frac{k_{isf}}{12}m\,L^2\,\ddot{\theta}_i
\end{aligned}
\end{equation}
Where $Fx_i$ and $Fy_i$ are the reaction forces from the revolute joints in the horizontal and vertical directions respectively, $k_{isf}$ is a unitless scaling factor to account for gearbox inertia.  $C_x$ and $C_y$ are environmental drag forces in $x$ and $y$ directions. Reaction forces and torques from the next distal link are absent for the last link and thus $Fx_{i+1}$, $Fy_{i+1}$ and $\tau_{i}$ are zero for the last link. For the stance phase the Cartesian positions ($x_{1-3}$, $y_{1-3}$) and speeds ($\dot{x}_{1-3}$, $\dot{y}_{1-3}$)  of links are expressed in terms of the generalized coordinates and speeds of the system state $\theta_{1-3}$ and $\dot{\theta}_{1-3}$, respectively. Here link one is the head link and link three is the posterior link. The system of nine equations can be obtained for  $\ddot{\theta}_{1-3},\,Fx_{1-3}$ and $Fy_{1-3}$ as a function of the system states ($\theta_{1-3}$ and $\dot{\theta}_{1-3}$) and input torques $\tau_{1-2}$. The final   governing equation can be written as, 

\begin{equation}
\label{ODE}
\mathbf{M}(\boldsymbol{\theta})
\begin{bmatrix}
\ddot{\boldsymbol{\theta}}\\
\mathbf{f}
\end{bmatrix}
=
\mathbf{C}(\boldsymbol{\theta})\,\dot{\boldsymbol{\theta}}^2
+\mathbf{G}(\boldsymbol{\theta},\dot{\boldsymbol{\theta}})
+\mathbf{U}\mathcal{T}
\end{equation}
where
\begin{equation*}
\begin{array}{cccc}
\boldsymbol\theta = 
\begin{bmatrix}
\theta_1 \\
\theta_2 \\
\theta_3 
\end{bmatrix}
&
\dot{\boldsymbol{\theta}}^2 = 
\begin{bmatrix}
{\dot \theta_1}^2\\ 
{\dot \theta_2}^2\\ 
{\dot \theta_3}^2\\
\end{bmatrix}
& 
\mathbf{f} =\begin{bmatrix}
Fx_{1}\\ 
Fy_{1}\\ 
Fx_{2}\\ 
Fy_{2}\\ 
Fx_{3}\\ 
Fy_{3}\\ 
\end{bmatrix}
& 
\mathcal{T} =\begin{bmatrix}
\tau_1\\
\tau_2\\
\end{bmatrix} 
\end{array}
\end{equation*}
while 
\[
Fy_1 > 0.
\]

Here, $\mathbf M$ is a $9 \times 9$ matrix as a function of the robot geometry. The matrix $\mathbf C$ is a $9 \times 3$ is a nonlinear function of position and accounts for the coriolis forces. The symbol $\dot{\boldsymbol \theta} ^2$ is a $3\times 1$ vector of the element-wise squares of the time derivatives of $\boldsymbol \theta$, and $\mathbf G$ is a $9\times1$ vector represents the effect of gravity and environmental forces  acting on the linkages. $\mathbf{U}$ is a $3 \times 2$ input matrix with $-1$ on the main diagonal and $1$ on the subdiagonal and zeros everywhere else. These equations apply while $Fy_1$ is positive.

 \subsection{Flight}
During the flight phase, the head is free to translate as well as rotate, and the dynamics of the serial linkage can be described with five position states and five velocity states instead of the previous three and three, respectively. Similarly, four reaction forces are required instead of the previous six. The coupled ODEs with nine equations and nine unknowns can be expressed as:

\begin{equation}
\label{ODEflight}
\mathbf{M}(\boldsymbol{\theta})
\begin{bmatrix}
\ddot{\boldsymbol{\theta}}\\
\ddot x\\
\ddot y\\
\mathbf{f}
\end{bmatrix}
=
\mathbf{C}(\boldsymbol{\theta})\,\dot{\boldsymbol{\theta}}^2
+\mathbf{G}(\boldsymbol{\theta},\dot{\boldsymbol{\theta}},\dot x,\dot y)
+\mathbf{U}\mathcal{T}
\end{equation}
where $\mathbf{f} =\begin{bmatrix}
Fx_{2}& 
Fy_{2}& 
Fx_{3}& 
Fy_{3}
\end{bmatrix}^T$ 

 and ($x$,$y$) is the Cartesian position of the first link. These equations apply while the head's vertical speed is positive in the inertial reference frame.
 
 \subsection{Transitions}
The conditions that result in transitions between the flight and stance phases are functions of the states. Our model assumes lift-off occurs when the $y$ component of the head reaction force becomes zero. 

When this occurs, the system states at the end of the stance phase are mapped into the initial stages of the flight phase according to the following transform,
\begin{align}
\label{stance2flight}
\begin{small}
 \begin{pmatrix}
 \theta_{1fi}\\
 \dot{\theta}_{1fi}\\
 \theta_{2fi}\\
 \dot{\theta}_{2fi}\\
 \theta_{3fi}\\
 \dot{\theta}_{3fi}\\
 x\\
 \dot{x}\\
 y\\
 \dot{y}\\
 \end{pmatrix}
 =
 \begin{pmatrix}
 1      & 0 & &\dots   & 0\\
 0      &\ddots& & &\\
 \vdots & &    & & \vdots\\
        &  &        & \ddots &0\\
   0    & \dots&    & 0& 1\\
 0 &  \dots & & &0\\
 0 & -\frac{L}{2}\,\text{s}(\theta_1) & 0 &\dots & 0\\
 0 &  \dots& & & 0\\
 0 & \frac{L}{2}\,\text{c}(\theta_1) & 0 & \dots & 0\\
 \end{pmatrix}
 \begin{pmatrix}
 \theta_{1s}\\
 \dot{\theta}_{1s}\\
 \theta_{2s}\\
 \dot{\theta}_{2s}\\
 \theta_{3s}\\
 \dot{\theta}_{3s}\\
 \end{pmatrix}
 +\begin{pmatrix}
 0\\
 0\\
 0\\
 0\\
 0\\
 0\\
 \frac{L}{2}\text{c}(\theta_1)\\
 0\\
 \frac{L}{2}\text{s}(\theta_1)\\
 0\\
 \end{pmatrix}
 \end{small}
\end{align}
where $c(\theta)$ and $s(\theta)$ stand for $\cos$ and $\sin$ respectively. In Equation \ref{stance2flight} the vector on the left-hand side is the initial conditions for the flight phase states and the vector containing $\theta_{1-3}$ and $\dot{\theta}_{1-3}$ on the right-hand side is the states at the end of the stance phase.

We base our model on the passive attachment directional adhesion mechanism  resembling geckos and cockroaches which achieve good performance in other dynamic climbing studies. Previous studies show that such a passive attachment  can be modeled by assuming the negative spine velocity defines the moment of sticking \cite{brown2019evidence}. As such, the exit condition for the flight phase is based on the relative velocity of the microspine array at the head of the serial linkage given by the following equation for the vertical velocity of the head: 
\begin{equation}
\label{flightPhaseExitCondition}
\dot{y}_{head} = \dot{y}+\frac{L}{2}\cos(\theta_1-\pi)\,\dot \theta_1 > 0
\end{equation}

In summary, climber dynamics are governed by Equation \ref{ODE} while the head force is positive; then, the stance states are mapped to the flight states by the transformation given in Equation \ref{stance2flight}. Finally, the climber motion is governed by Equation \ref{ODEflight} until the vertical head velocity reduces to zero when the flight phase ends. The resulting hybrid dynamic motion governed by the stance and flight phases constitutes a single jump.

\subsection{Control Policy}
Hirose's early snake robot control policies prescribed feedforward or open-loop joint angle profiles generating snake-like motion he called serpanoid curves \cite{hirose1993biologically}. Later, Hatton et al \cite{hatton2010generating} developed methods for generating joint space trajectories or gaits that generate desired locomotion shapes using ``keyframes". Hatton's process produced the helix shape that enables tree climbing and other scrpted gaits. Later improvements on this approach allowed for climbing on changing pole diameters and delicate structures \cite{rollinson2013gait}. In the dynamic climbing and running community, the earliest successes were demonstrated with a feedforward control policy\cite{saranli2001rhex,spenko2008biologically}. In this work, we first explore a gait with a feedforward control policy of the joints angle that mimics a lamprey's neural 'clock' \cite{koditschek2004mechanical}. Our control policy parameterizes Trident Climbers' shape changes by specifying a compression speed, extension speed, compression magnitude, and rest time. 
For the three-link model employed here, we fully specify a control policy by describing the commanded joint speed in compression $\omega_C$, the commanded joint speed in extension $\omega_E$, an amount of compression $\gamma_{max}$ and a rest time $t_r$. This study reduces the parameter space by setting the second joint angle to always be $180^o$ out of phase with the first.

   \begin{figure}[h]
      \centering
      \includegraphics[width = \columnwidth]{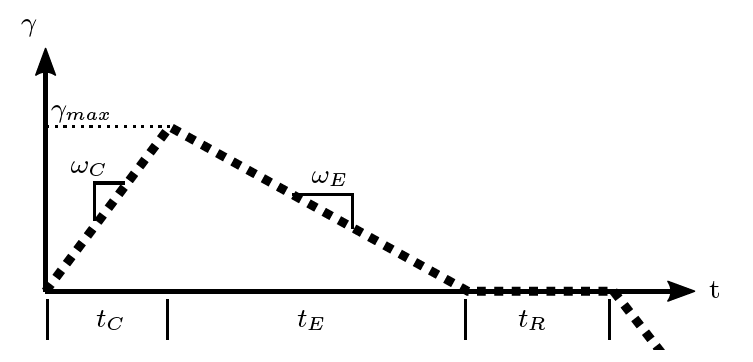}
      \caption{Triangular-shaped joint actuation pulse. These relative angles are the joint angles in Trident's joint space.}
      \label{actuationPulseFig}
   \end{figure}

Given the desired joint angle profile in figure \ref{actuationPulseFig}, we use a proportional derivative feedback control law based on the actual relative joint angles and compute the joint actuation torques [$\tau_{c1}\, \tau_{c2}$] as 
\begin{equation}
\label{pdlaw}
\begin{split}
\begin{pmatrix}
\tau_{c1}\\
\tau_{c2}
\end{pmatrix}
=
& K_p
\begin{pmatrix}
\phantom{-}\gamma_c-\gamma_1\\
          -\gamma_c-\gamma_2
\end{pmatrix}
+
K_d\,
\begin{pmatrix}
\phantom{-}\omega_{C/E}-\dot{\gamma}_1\\
          -\omega_{C/E}-\dot{\gamma}_2
\end{pmatrix}
\end{split}
\end{equation}
The calculated joint torques  $\tau_{c1}$ and $\tau_{c2}$ are subjected to the following motor model that limits the commanded joint torque to a saturation limit given by the linear motor model,
\begin{equation*}
    \tau_{max} = \left\{\begin{array}{llcc}
        \frac{v}{v_m}\tau_{ST} ,  
            &               \hspace{-0mm} -\infty   <    \dot \gamma \leq  0\\
         -\frac{\tau_{ST}}{\omega_{NL}}\,\dot \gamma+\frac{v}{v_m}\tau_{ST}, 
            &                         0 <    \dot \gamma \leq  \frac{v}{v_m}\omega_{NL}\\
        0,                                             
            & \frac{v}{v_m}\omega_{NL}  < \dot \gamma        <  \infty
        \end{array}\right.
  \end{equation*}
\begin{equation*}
    \hspace{+6mm}\tau_{min} = \left\{\begin{array}{llcc}
        0,                                                                   
            &                -\infty   <    \dot \gamma \leq -\frac{v}{v_m}\omega_{NL}\\
        -\frac{\tau_{ST}}{\omega_{NL}}\,\dot \gamma-\frac{v}{v_m}\tau_{ST} , 
            &  -\frac{v}{v_m}\omega_{NL} <    \dot \gamma \leq  0\\
        -\frac{v}{v_m}\tau_{ST},                                             
            &      \hspace{0mm}       0            < \dot \gamma <  \infty
        \end{array}\right.
\end{equation*}
where the parameters of the motor model are given in table \ref{tab:JumperParams}. Other motor and control policy parameters were discovered through the numerical optimization in section \ref{NumStudy}.

 \section{Numerical Study}
 \label{NumStudy}
  The hybrid dynamic system composed of equations \ref{ODE} and \ref{ODEflight}  was solved using Matlab's ODE45 algorithm, and the mathematical transformation in equation \ref{stance2flight} was implemented using the events parameter. This numerical study was first used to find an optimal robot morphology and controller. The optimization problem is stated in equation \ref{coOptimization}. Here we consider the jump height as a function of the design vector $X$.   The design Vector $X$ includes Trident Climber's link length $L$, joint trajectory parameters: $\omega_C,\, \omega_E \text{ and } \gamma_{max} $ and PD control gains $K_p\text{ and } K_d$.  

\begin{equation}
\label{coOptimization}
\begin{aligned}
& \underset{X}{\text{minimize}}
& -\Delta\,y(X) \\
\end{aligned}
\end{equation}
where
\[
X =
\begin{bmatrix}
L\\
\gamma_{max}\\
\omega_{C}\\
\omega_{E}\\
K_p\\
K_d\\
\end{bmatrix}
\]
For a given stall torque and no-load speed associated with three different available motors, a single shooting approach utilizing Matlab's \textit{fminsearch} was used to find a locally optimal design vector $X$.  The design parameters for three available motors as well as the predicted jump height are reported in Table \ref{tab:optimization outps}. The 75:1 reduction ratio produces superior performance, and the two equivalent power motors are shown with corresponding optimized design vectors for sensitivity comparison. These simulations were performed drag free ($C_x=C_y=0$) and thus the resulting jump heights are optimistic relative to what will be presented in the experimental section \ref{results}. 

A numerical parameter variational study was also conducted in order to visualize the climbing performance of this jumper, we systematically explore a wide range of the compression speed $\omega_C$ and the maximum actuation joint angle $\gamma_{max}$. Other fixed parameters of this test are given in table \ref{tab:JumperParams}. The contour plots of vertical and lateral displacement, as well as efficiency, are shown as contour plots in section \ref{results}. The parametric study is done for $5^o$ increments of the joint angle and $70$ deg/s increments of the compression speed, resulting in 880 data points. Each individual simulation result is shown as a gray dot in section \ref{results}.

\begin{table}[t]
    \begin{center}
 \begin{tabular}{|c| c c c c c c |c|}
  \hline
 \thead{motor                   } &
 \thead{$L$               \\ $[mm]$               } &
 \thead{$\omega_{C}$    \\ $[\frac{deg}{s}]$  } &
 \thead{$\omega_{E}$    \\ $[\frac{deg}{s}]$  } &
 \thead{$\gamma_{max}$  \\ $[\frac{deg}{s}]$  } &
 \thead{$K_p$                         } &
 \thead{$K_d$                         } &
 \thead{$\Delta y$       \\ $[mm]$         } \\
 \hline\hline
 50:1  & 89     & 718                  & 682                    & $125^o$         &1.6 &.02      & 55\\ [1ex]
 \hline
 75:1  & 182   & 1580                  & 900                    &$140^o$         &1.0   &0.1          &122\\ [1ex]
 \hline
 100:1  & 141  & 1324                  & 718                    &$132^o$        &1.1     &.05             & 93\\ [1ex]
 \hline
\end{tabular}
\end{center}
    \caption{Design vectors that maximize jumping height for three different motor choices. Of the motors considered, these gearing choices bracket the optimal choice.}
    \label{tab:optimization outps}
\end{table}

\section{Robot Platform -- Trident Climber}
\label{sec:RobotPlatform}
\begin{figure}[h]
      \centering
      \includegraphics{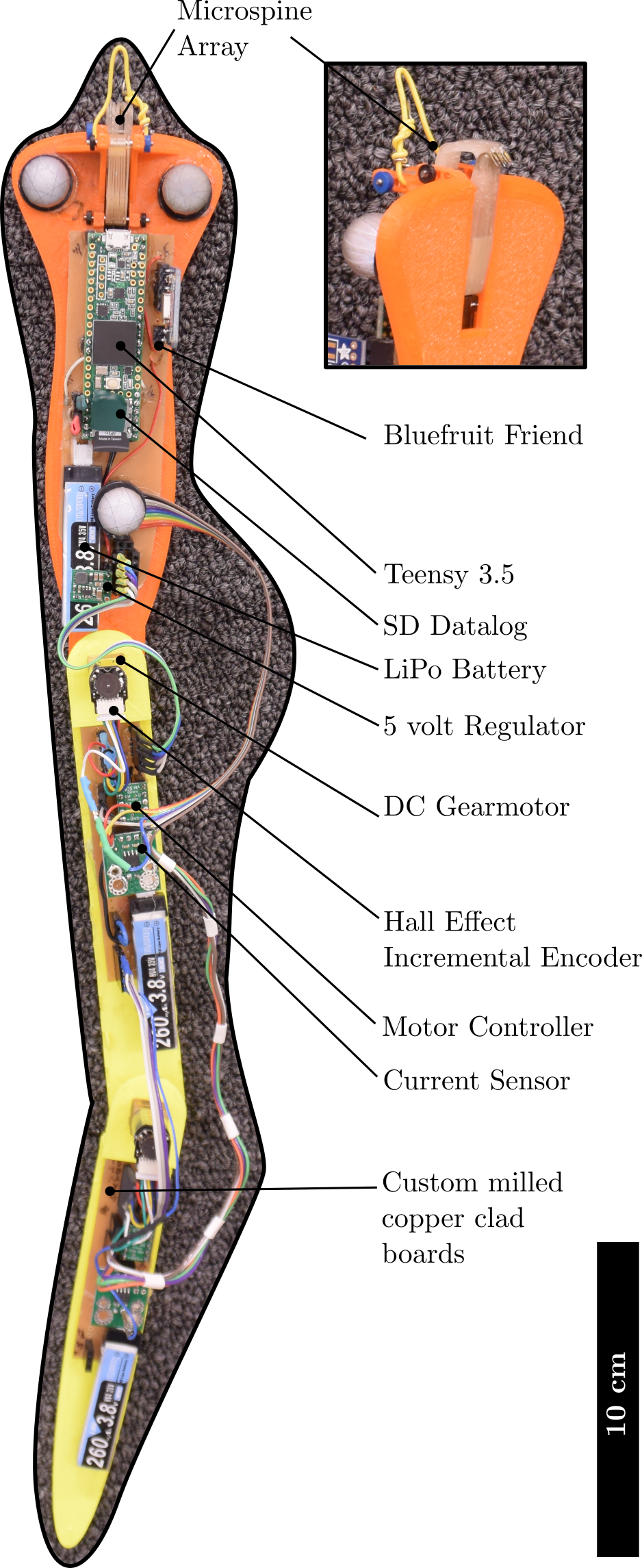}
      \caption{Trident Climber Hardware Diagram}
      \label{fig:HardwareDiagram}
   \end{figure}

The robotic platform Trident Climber, shown in figure \ref{fig:HardwareDiagram}, was constructed in order to test and validate the three-link model and demonstrate a lamprey-inspired gait in hardware. Trident Climber is composed of three 3D-printed ABS segments connected and driven by 2 DC gearmotors. Link lengths were chosen to optimize the jump height as described in section \ref{NumStudy}.

\begin{table}[b]
    \centering
\caption{Jumper Model Parameters}
\label{tab:JumperParams}
 \begin{tabular}{||c c c c||}
 \hline
 Parameter & Symbol & Value & Units \\
 \hline\hline
 Mass per link & m & 54 & $g$ \\ [1ex]
 \hline
  Acceleration due to gravity & g & 9.81 & $ms^{-2}$ \\ [1ex]
 \hline
 Climbing wall angle & $\alpha$ & $20$ & deg \\ [1ex]
 \hline
 Link length & $L$ & 182 & $mm$ \\[1ex]
 \hline
 Coulomb drag force in $Y$ per link & $C_y$ & 0.45 & $N$ \\[1ex]
 \hline
 Coulomb drag force in $X$ per link & $C_x$ & 0.08 & $N$ \\[1ex]
 \hline
 Rotational Inertial Scale Factor\footnote{$K_{isf}$ was experimentally determined by tuning model to match the experimental data} & $k_{isf}$ & 2 & - \\
 \hline
  Motor voltage   & $v$ & 11 & Volts \\ [1ex]
 \hline
 Motor spec voltage   & $v_m$ & 6 & Volts \\ [1ex]
 \hline
 Motor no-load speed & $\omega_{NL}$ & 410 & RPM\\[1ex]
 \hline
 Stall torque & $\tau_{ST}$ & 127 & $N \,mm$ \\[1ex]
 \hline
 \hline
 \end{tabular}
\end{table}

 These optimizations suggest that Pololu's dual shaft High Power Micro Metal Gear Motor with a 75:1 gear head (part no: 2361) and a 182 mm link length are the best hardware choices. A hall effect magnetic incremental encoder is attached to the motor rotor output shaft, and the output of the gearhead is connected to an aluminum coupler fixed to the bottom of the adjacent forward link. The combination 75:1 speed-reducing gearhead and the twelve count per revolution encoder provide relative joint angle feedback at about 910 counts per revolution. Both joint's relative joint angle $\gamma$ are capable of $\pm 90^o$ of rotation measured from the previous link's centerline.
 
  Trident Climber is equipped with eight microspines \cite{asbeck2006scaling} forming an array at the tip of its head analogous to  \textit{L. tridentata's} oral disk and buccal funnel. Microspines are built by a shaped deposition manufacturing \cite{weiss1997shape} technique of sequential material removal and addition. In this way, a composite array of compliant hooks is made that is highly effective at engaging surface asperities in a climbing substrate\cite{spenko2008biologically}.
  
  Communications are facilitated by an Adafruit Bluefruit LE SPI Friend (part number  2633).  Trident Climber is controlled by a PJRC Teensy 3.5 microcontroller. The Teensy logs data as well as generating the joint angle profiles, reading incremental encoders, and sending PWM control signals to the motor drivers based on a 1kHz digital implementation of the control policy in Equation \ref{pdlaw}.
  
 System power is supplied by three LiPo batteries connected in series, forming a nominal 11.5 volt bus. This voltage is regulated to five volts for the microcontroller, encoders, and logic signal level supply and delivered unregulated to the driving motors. Trident Climber is equipped with a Pololu current sensor (part no. 4041) for logging motor current. Trident Climber equipped with batteries and microspines weighs 162 gm.

\section{Experimental Procedure}
\label{ExperimentalSection}

   \begin{figure}[t]
      \centering
      \includegraphics{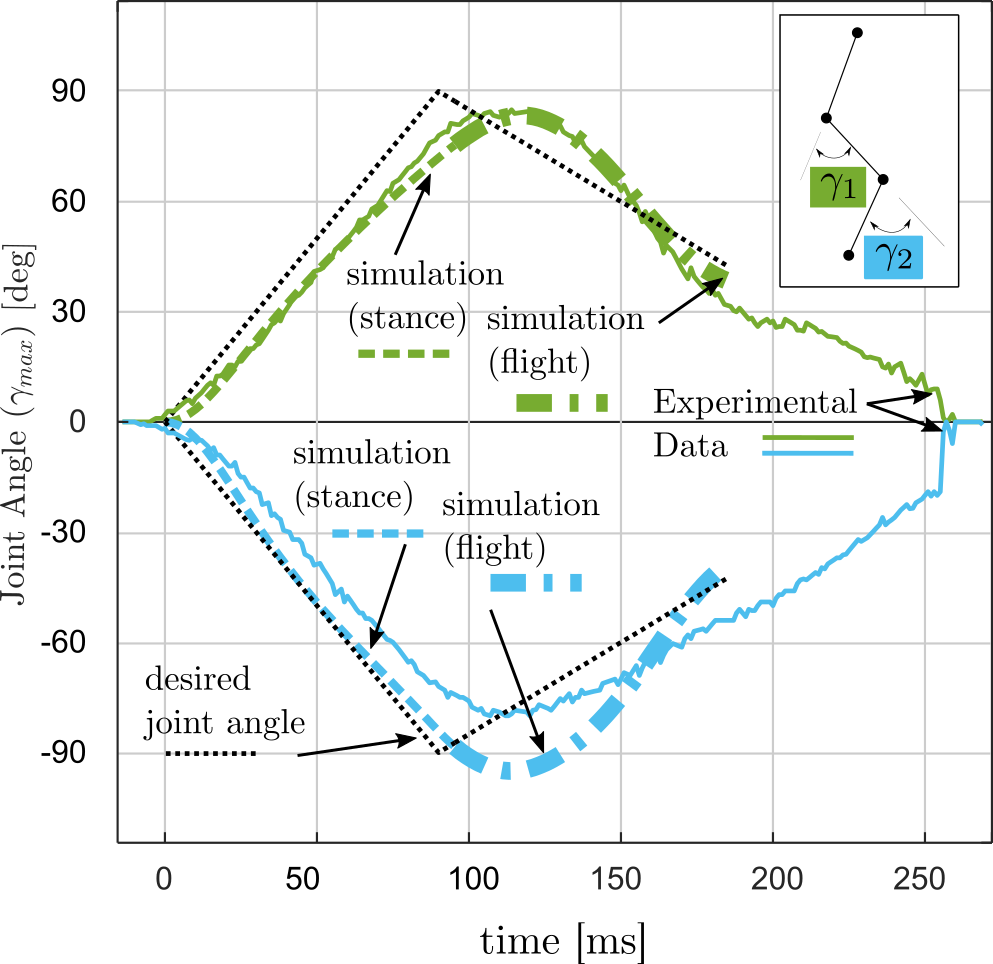}
      \caption{Comparison of experimental data and simulation shape kinematics in the shape space  during one jump at compression speed of 1000 deg/s, extension speed of 500 deg/s, and maximum relative joint angle of $90^o$. The first joint angle defining the head-to-middle section angle $\gamma_1$ is shown in green achieving approximately $90^o$ and the second distal joint angle $\gamma_2$ defining the angle between the middle and tail pieces is shown achieving approximately $-90^o$. Here the experimental data trace is the average for 29 jumps.}
      \label{fig:shapeVars}
   \end{figure}

 \subsection{Model Calibration}

 Figure \ref{fig:HardwareDiagram} shows reflective markers for logging the head link's position. A Vicon motion capture system with four calibrated cameras is employed to record the position data at 100 Hz. Preliminary testing revealed that drag plays a significant role in the dynamics of this form of climbing at this mass scale. In order to account for drag and spine release cost, Trident Climber was placed on its climbing surface and a pulley system was used to add weight incrementally to determine drag force in the $y$ direction. The value of this drag force is reported on a per link basis in table \ref{tab:JumperParams} as $C_y$. Lateral drag was similarly measured and reported in table \ref{tab:JumperParams} as $C_x$. Here we assume that lateral motion in each link's local coordinate frame, can be reasonably mapped to the global $x$ direction and local link vertical motion is primarily in the $y$. In table \ref{tab:JumperParams} it can be seen that $C_y$ is much greater than $C_x$,  0.48 vs 0.08 respectively. This disparity is due to the force required to make the microspines release. This spine release cost is not present for motion in the lateral direction.  Moreover, the scaling factor $k_{isf}$ to account for the gearbox reflected inertia was found to be $2.0$.
 \subsection{Hardware Experiments}
\begin{figure}[t]
      \centering
      \includegraphics{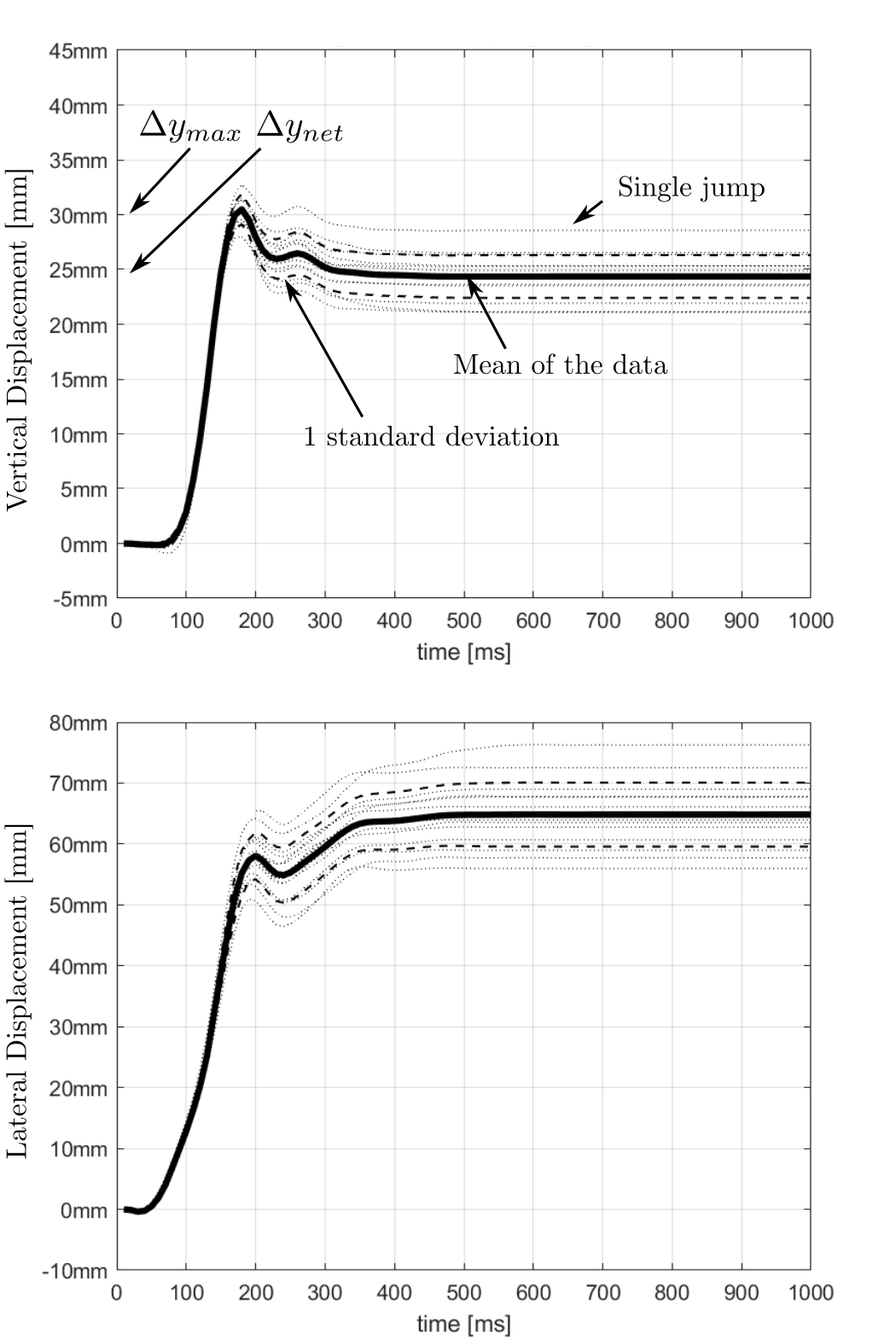}
      \caption{Example of post-processed Vicon Data from a single jumping experiment at compression speed $\omega_C$ of 1000 deg/s and maximum relative joint angle of $\gamma_{max}$of $60^o$. This data point is \ref{fig:experimentalSimulationTile}(c,f,i) at the $60^o$ operating point. Data is sampled at 100 samples per second and binned allowing the mean and standard deviation to be found for each sample time after the beginning of the jump.}
      \label{fig:ViconExample}
   \end{figure}

 In this study, we consider climbing behaviors that allow vertical and lateral motion. We quantify the performance of the climbing by evaluating the net vertical displacement, the net lateral displacement, and energetic efficiency. For each experiment the robot was commanded to jump 29 times, 15 right and 14 left. Experimental data of each jump was post-processed such that the vertical and lateral displacement was evaluated separately. The time-series data from the Vicon and the onboard joint encoder and current log were binned such that the mean and standard deviation of the peak jump could be determined for each experiment. Sample experimental results of this internal shape data for a 29 jump experiment is compared to the three-link model numerical results in Fig. \ref{fig:shapeVars}. Here we see the correspondence between the simulation shape variables ($\gamma_1, \gamma_2$) and the shape variables as measured by Trident's incremental encoders. Fig. \ref{fig:shapeVars} also depicts the simulated transition from stance to flight phases, shown with the dashed and dotted-dashed lines, respectively. An example of global reference frame data from the Vicon system is shown in Fig. \ref{fig:ViconExample}. This plot gives the time histories for Trident's binned joint angles during one experiment.
 
 Efficiency is computed by comparing the ratio of work done by the motors at the output shaft to the increase in potential energy of Trident Climber. The work done by the motor is calculated by time integrating the instantaneous power of each motor. This is done using the measurements from the joint encoder and motor current according to Equation \ref{eq:efficiency}. 
 \begin{equation}
 \label{eq:efficiency}
 \eta=
 \frac{m\,g\,\Delta y}{\int_0^{t_{flight}}P_{motor}\,dt}=
 \frac{m\,g\,\Delta y}{k_T\int_0^{t_{flight}}\dot{\gamma}_1\,i_1
 +\dot{\gamma}_2\,i_2
 \,dt}
 \end{equation}
 where $\dot \gamma$ is the motor speed and $k_T$ is the motor torque constant.
\section{Finding parameters for peak climbing -- Simulation and Experimental Results}
\label{results}

\subsection{Parameter Variation Study}

\begin{figure*}
    \centering
    \includegraphics[width=1.85\columnwidth]{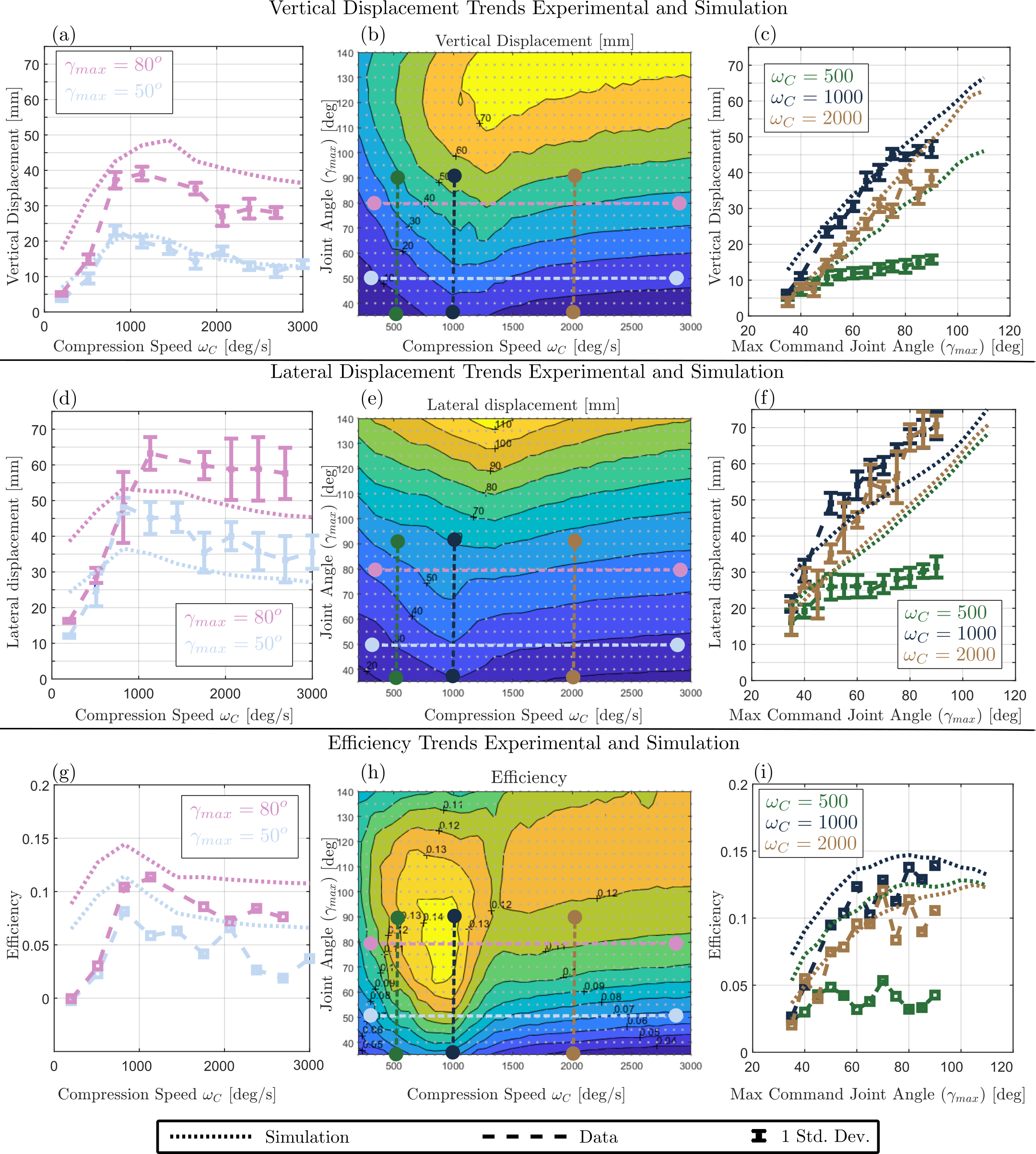}
    \caption{Results of 57 jumping experiments compared against simulation showing results of varying the maximum commanded relative joint angle $\gamma_{max}$ and the command compression speed $\omega_C$. From top to bottom, the Rows of the figure represent vertical displacement, Lateral displacement, and locomotion efficiency. Left-most column depicts the effect of compression speed $\omega_C$ by holding $\gamma_{max}$ constant. Simulation is also compared against experimental data. The middle column shows simulated contour plots with color-coded dashed lines showing slice locations. The rightmost column plots the effect of $\gamma_{max}$ on the performance metrics of interest for several compression speed $\omega_C$ values. Gray dots represent a simulated data point.}
        \label{fig:experimentalSimulationTile}
\end{figure*}

 The outside columns of figure \ref{fig:experimentalSimulationTile} show comparisons of the experimental data with error bars reflecting one standard deviation of data as in figure \ref{fig:ViconExample}. For instance, the data represented in figure \ref{fig:ViconExample} is a single data point in the cyan trace at $\gamma_{max} = 60^o$ in subfigure \ref{fig:experimentalSimulationTile}(c). 
 
  Fig. \ref{fig:experimentalSimulationTile}(a) shows the effect of compression speed on vertical displacement and each data set (single color) is a horizontal cross-section of Fig. \ref{fig:experimentalSimulationTile}(b). The jump height is positively correlated with the compression speed until around $\omega_c \approx 1000\, deg/s$, when it begins to decrease with further increase of $\omega_c$.  Experimental observations and simulation results both display the same location of peak jump height and a slight dependence on the max joint angle $\gamma_{max}$. However, the simulation model overestimates the jump height by about three millimeters.

 Fig. \ref{fig:experimentalSimulationTile}(c) shows the effect of the maximum commanded joint angle on the vertical displacement of a single jump and each curve is a vertical cross-section of Fig. \ref{fig:experimentalSimulationTile}(b). We observe that within the joint angle constraint of Trident Climber, vertical displacement is positively and linearly correlated with the max joint angle. Similar to what is observed in Fig. \ref{fig:experimentalSimulationTile}(a) the model over-predicts jump height by about three millimeters for most cases, except for low compression speeds. Peak vertical climbing displacement of 47mm $\pm5$ occurs for a commanded compression speed of $\approx 1000 deg/s$ and $\gamma_{max}$ of $90^o$.

 Fig. \ref{fig:experimentalSimulationTile}(d-f) shows the lateral displacement of a single jump. Similar to vertical jump displacement, there is a peak compression speed, $\approx 1000$ deg/s, with slight dependence on joint angle. Also, the lateral displacement has a positive linear correlation with joint angle. Unlike vertical displacement, the model underpredicts the lateral distance by about 10 to 15 millimeters in most cases.
  
 Efficiency is shown in Fig. \ref{fig:experimentalSimulationTile}(g-i). In \ref{fig:experimentalSimulationTile}(h) we observe that a peak efficiency of $14\%$ occurs at $\gamma_{max} = 90^o$ and $\omega_C = 1000$ deg/s.  Fig. \ref{fig:experimentalSimulationTile}(i) reports efficiency based on the $\Delta y_{max}$ (see fig. \ref{fig:ViconExample}.) As such, this number is the same output as the Numerical study. 
 In order to make a fair comparison of Trident's single jump data to continuous climbing and to other robots we consider the net vertical displacement (see fig. \ref{fig:ViconExample}.) Trident Climber slips some distance, typically about 5 mm, as its microspines engage the carpet. Consequently, the $14\%$ reported is the max efficiency that can be attained if the spines attach immediately at the apex. As such, the net efficiency for the peak efficiency operating point is $12\%$ or a specific resistance of 8.3.

\section{Locomotion Strategies}
\label{sec:LocomotionStrategies}

   \begin{figure}[h]
      \centering
      \includegraphics[width = \columnwidth]{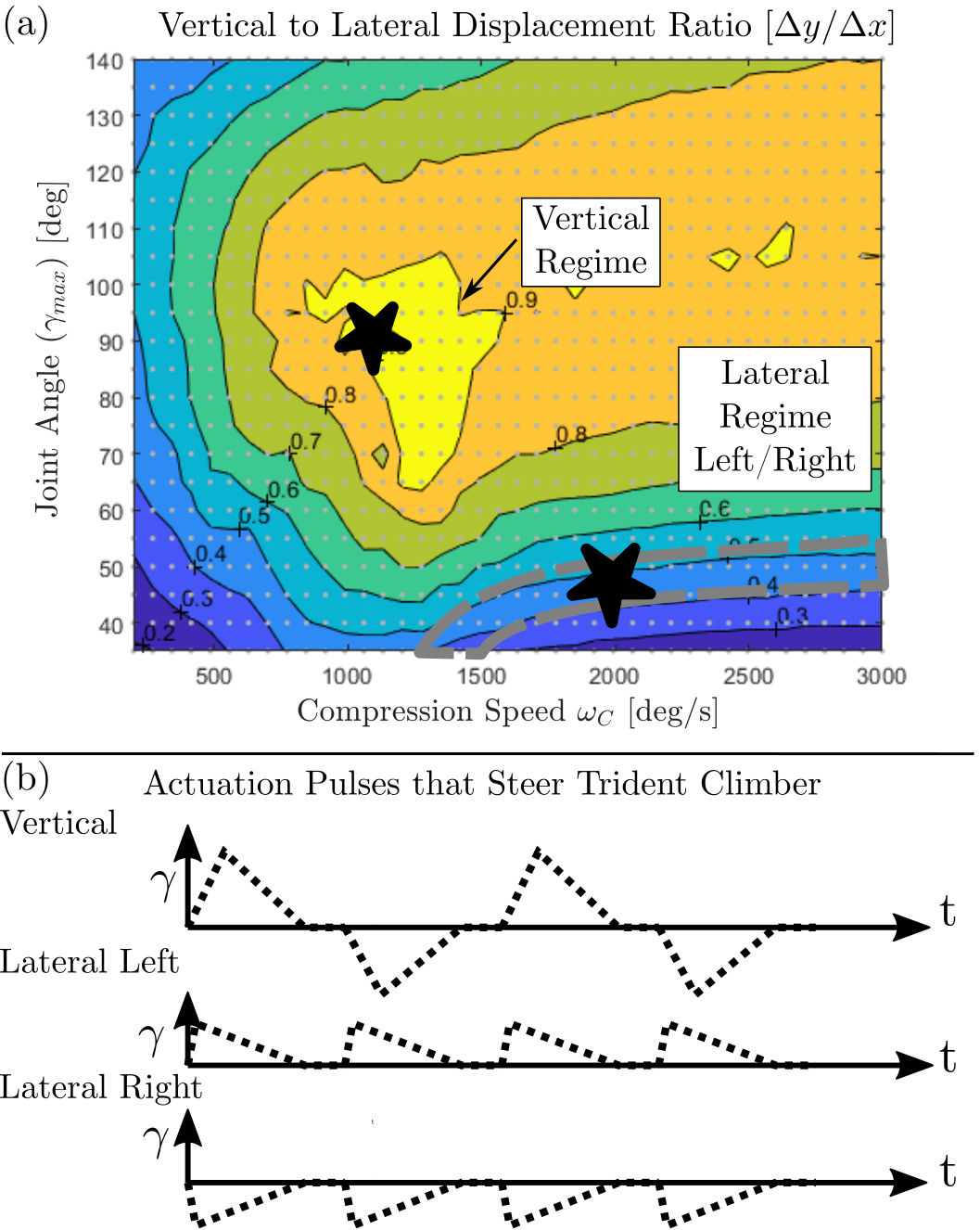}
     \caption{a.) Contour plot showing the vertical-to-lateral displacement ratio in the control parameter space. b.) Feedforward joint angle trajectories that cause vertical right and left locomotion. Note that Fig. \ref{fig:experimentalSimulationTile} vertical displacement is at peak. there is a displacement cost for reattaching on the order five to 10  mm thus $\gamma_{max}$ on the order of 50$^o$ results in negligible net vertical progress.}
      \label{fig:SteeringActuation}
   \end{figure}

   \begin{figure}[h]
      \centering
      \includegraphics[width = \columnwidth]{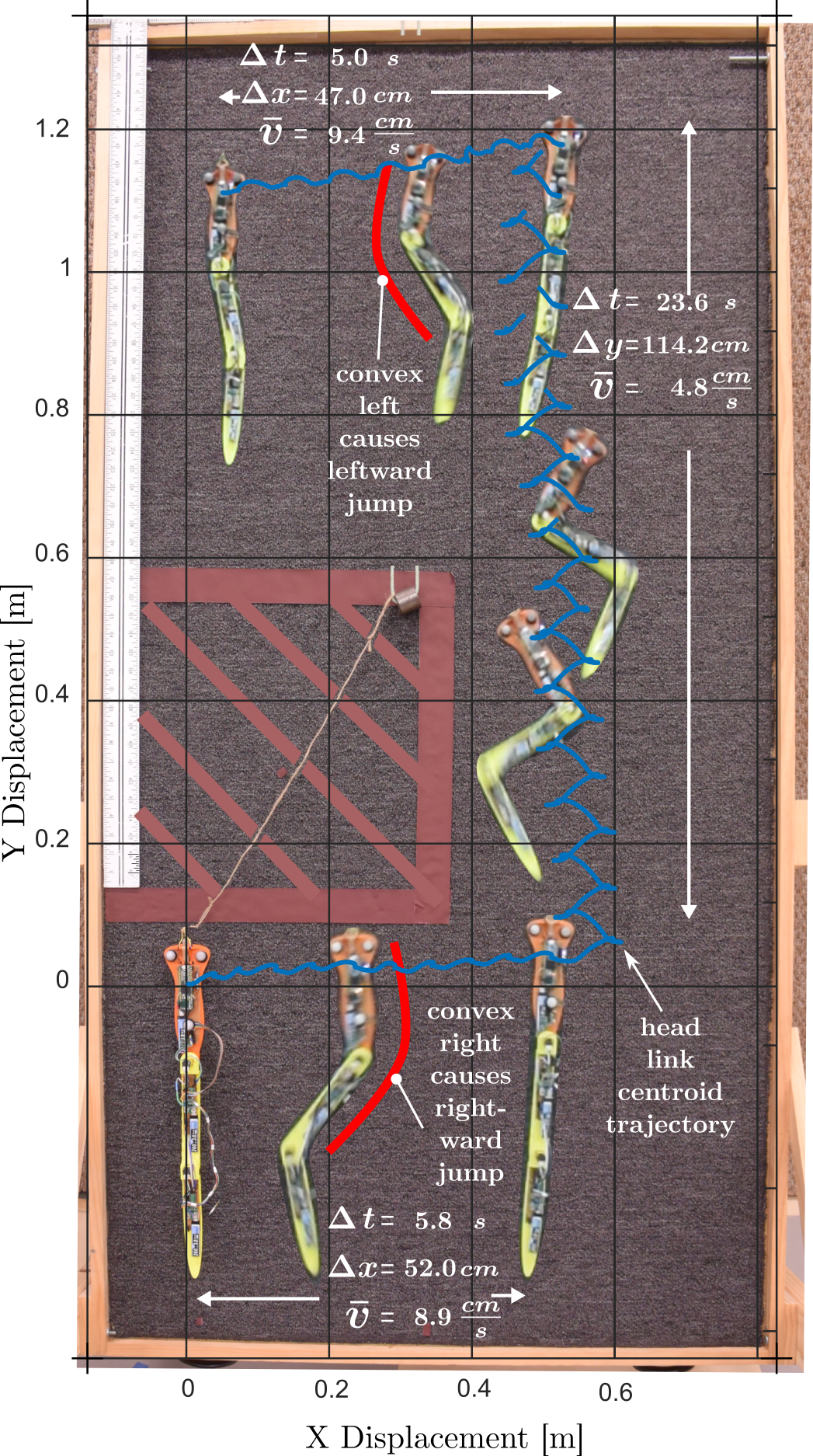}
      \caption{Steering trajectory for Trident Climber. Trident Climber steers around an obstacle by first applying 10 lateral left actuation pulses as depicted in Fig. \ref{fig:SteeringActuation}(b) Lateral Right with a $\gamma_{max}$ of 50$^o$ and compression speed $\omega_C$ of -2000 deg/s. Similarly, vertical climbing is achieved by actuating as in Fig. \ref{fig:SteeringActuation}(b) Vertical with 31 alternating sign compressions with $\gamma_{max}$ of 90$^o$ and $\omega_C$ of 1000 deg/s. Finally, Trident Climber traverses left with 10 pulses the same as vertical right but with positive compressions resulting in leftward jumps. Note that positive compression speeds result in the top two links convex left and result in left jumps; vice versa for negative compression speeds.}
      \label{fig:steering}
   \end{figure}
\subsection{Steering}
After determining the control parameter space that relates to climbing performance, we consider steering in the context of vertical climbing. Figure \ref{fig:SteeringActuation} is generated by computing the vertical-to-lateral displacement ratio for the parameter variational study. This plot reveals how the actuation parameters ($\gamma_{max}$, $\omega_C$) affect steering on Trident.  The actuation in the stared regimes can be used to steer Trident Climber, resulting in net motion that is either vertical or mostly lateral. The yellow contour labeled ``Vertical Regime" represents strides that have the largest vertical component. So if vertical progress is desired, and high lateral deviation is acceptable, the ``Vertical Regime" is the optimal operating point. This vertical jumping regime is characterized by a large joint angle $90^o$ and the commanded compression speed is appropriately matched to the characteristics of the motor ($\approx 1000$ deg/s for our motor).  To achieve mostly net vertical locomotion the actuation ($\omega_C$) must alternate sign since consecutive steps alternate displacement in the lateral direction. As such, two jump steps form a stride with a $y$ displacement of two times the vertical step displacements reported in Fig. \ref{fig:experimentalSimulationTile}.

 If, on the other hand, the objective is to achieve lateral motion, then Trident can be actuated in the region labeled as ``Lateral Regime Left/Right." Here the jumping is characterized by a low joint angle and a compression speed slightly faster than what is optimal for vertical climbing. For the marked case, the compression speed is $\approx 2000$ deg/s. Considering that the entire blue contour with a ratio between 0.3 and 0.4 has a similar displacement ratio, priority is given to the higher compression speed since it results in faster cycle speeds. However, considerations other than speed might lead one to choose the regime on the left side of the blue contour. As shown in Fig. \ref{fig:experimentalSimulationTile}(c), the peak vertical displacement for this particular case $(\omega_C=2000\, deg/s\,,\, \gamma_{max}=50^o)$ is about 10 mm, accounting for spine engagement displacement cost results in close to zero net lateral displacement and 25 mm lateral jumps.
 
Another key result is that the body shape in the stance phase predicts the jump direction. By the convention in Fig. \ref{jumperFBD}, positive compression speeds on the first motor $\dot{\gamma}_1$ result in shapes where the first two links are deflected with their convex sides pointing left and as a result, the net lateral motion is to the left. For example, the jumper in Fig. \ref{jumperFBD} is poised to jump left. Conversely, negative compression speeds, result in rightward lateral jumps. 

By incorporating insights about locomotion direction and the ratio of vertical to lateral displacement into our feedforward control scheme, we were able to pick a proper pulse profile, similar to the one depicted in Fig. \ref{fig:SteeringActuation}(b) to steer Trident Climber. Note that vertical locomotion utilizes a pulse with compression speed of alternating sign. The lateral motion utilizes small, fast pulses with a compression speed corresponding to the desired lateral displacement Fig. \ref{fig:steering} shows Trident Climber demonstrating the utility of these actuation pulses for steering around an obstacle. 

\subsection{Mechanism of Performance}
 The principle of impulse and momentum explains the positive correlation between max joint angle, vertical jump, and lateral motion. Impulse and momentum theory states that the change in the vertical velocity of Trident is proportional to the integral of the vertical head reaction force with respect to time. 
\begin{equation}
\label{eq:imp_mom}
\Delta\,\vec{V} \propto \int \vec{F}(t) \,dt
\end{equation}
Athletic jumps thoroughly saturate the motor and limit the head reaction force. Considering equation \ref{eq:imp_mom} and noting that though the value of $\vec{F}(t)$ is invariant due to saturation at any $t$, increasing the duration of the force profile will still increase the impulse and the vertical velocity. Conversely, When commanded speeds are outside the motor's limit, the head reaction force has a much shorter time interval to develop a vertical impulse. For this reason, peak performance observed in terms of height, efficiency, and to a lesser degree, lateral travel is due to a good matching of command pulse to motor power output and link length. Furthermore if maximizing vertical step displacement is the primary goal, consideration should be given to maximizing the range of joint motion.

The experimental results and the model predictions agree well except for at low compression speeds. The poor correlation between the experiments and the model in this region is presumably because the energetic cost of detachment is significant compared to the kinetic energy gained during the stance phase; therefore, the revolute joint assumption used to  model the head attachment does not represent the dynamics of these cases.

Generally speaking, the stance-to-flight transition occurs when commanded compression ends and commanded extension begins. In other words, the stance phase is roughly equivalent to when the body is compressing and a flight phase begins when the actual max joint angle is reached and the extension phase begins.

\subsection{Comparison to Biological Exemplar}
Kemp and Mosser documented that \textit{L. tridentata} climbs with a mean stride length of 3.59 cm \cite{kemp2009linking}. Trident Climber achieves a step height of $4.1\pm 0.25$ cm in the vertical climbing regime, which is about $14\%$ better than the mean cycle step height published for \tridentata.   

Considering  Fig. \ref{gammaDef}, and applying the law of cosines to the triangles generated by shortening the overall length of the three links we can write down the relationship between sinuosity $\Omega$ and $\gamma_{max}$ given as
\begin{equation}
\label{gammaVsOmega}
\gamma_{max}(\Omega)=\pi-\arccos\left\{\frac{5}{4}-\frac{9\Omega^2}{4}\right\} 
\end{equation}

Kemp et al publish that {\tridentata} climb with a sinuosity of 88\%. Equation \ref{gammaVsOmega} gives  $\gamma_{max}(0.88) \approx 60^o$ suggesting that Trident Climber's compression is comparable to observed compression in \textit{ L. tridentata}. Previously, Zhu and Mosser suggested that lampreys climb in such a way as to optimize their climbing efficiency, and as can be seen in Fig. \ref{fig:experimentalSimulationTile}(h), $\gamma_{max} = 60^o$  is indeed close to an efficiency peak for Trident Climber.

When we compare our compression and extension speeds of Trident Climber with previously published data from \tridentata, 
Kemp et al \cite{kemp2009linking} show the compression time of {\tridentata} is about 160 ms and this range is similar when looking at other videos of Lampreys climbing. Keeping the Froude number constant and scaling to Trident Climbers length scale, this corresponds to 450 deg/s compression speed 
\cite{miller2015dynamic}. Figure \ref{fig:experimentalSimulationTile}(h) shows the efficiency of this location ($\gamma_{max}=60^o, \omega_C=450\,deg/s$) is approximately 10\%. The efficiency of this operating point is within 25\% of the model's peak. This minor discrepancy is perhaps due to \textit{L. tridentata} being force limited in its stance phase, muscles or oral disc holding power, or attachment dynamics.

We could not compare the lateral displacement performance between Trident Climber and its biological counterpart as, to our knowledge, no prior work has quantified the lateral motion by {\tridentata } while climbing.

\section{Conclusion and future work}
\label{sec:conclusions}
We demonstrate a lamprey-inspired climbing gait using a simple control policy  for climbing near-vertical surfaces both in simulation and in hardware. We envision that this attachment and body actuation profile will allow other snake robots to climb on certain steep terrain with dynamically similar performance.

We use a Newton-Euler model for a three-link pendulum to design a lamprey-inspired robot and validate the model with experimental data. We find that the resulting motion gives qualitatively similar motion to the animal exemplar \textit{L. tridentata} and produces an efficiency peak performance for body actuation geometry very close to the animal and previous continuum models. These numerical and experimental findings show that a three-link pendulum can function as a reduced-order model for lamprey-inspired climbing.

Many bio-inspired robots fail to perform as fast and agile as their animal counterparts. Trident climber achieved peak 4.7 cm strides for single jump tests and continual climbing speeds of 4.8 cm/s 

Both the single jump and continual climbing performance of Trident exceeds the mean performance of the animal exemplar climbing at similar frequencies even though Trident is approximately $75\%$ of the mean length of animals studied \cite{kemp2009linking}.

Trident achieves a specific resistance of 8 whereas other dynamic climbing robots have achieved as low as 5 and better \cite{provancher2010rocr}. Such a reduction in efficiency tends to make lamprey-inspired locomotion less compelling for only a climbing task. However, given that this 60\% reduction in efficiency comes with a long slender body morphology with the various strengths of its body's small cross-sectional area and is known to be able to swim \cite{porez2014improved}, a snake robot morphology should be given strong consideration for a multi-modal robot that must swim and climb. This inclination to long slender morphology is bolstered by noting that the joint actuation is essentially the same control policy applied in a pulsed way for climbing and a continuous way for swimming.

In terms of future work, Trident requires a head design that moves in the dorso-ventral plane to facilitate attachment when climbing wall angles beyond 70$^o$. Additionally, a suction-based design could allow for climbing on smooth surfaces such as ship hulls. Moreover, modeling and experimentation show that the stride length of lamprey-inspired climbing is positively correlated up to a $90^o$ joint angle. The simulations indicate that this trend continues up to $\gamma_{max} = 140^o$; therefore, future lamprey-inspired climbers should consider ways to achieve more flexibility for better performance. The simple control policy employed in this study demonstrates the stability of this locomotion, but utilizing trajectory optimization would allow for faster, more efficient gaits. Based on this modeling, we hypothesize that Lampreys coordinate the buccal funnel such that it releases at its body's max compression and reactivates at the apex of head flight. Some experiments with {\tridentata} with synchronized video and force plate data are needed to confirm this hypothesis.

\section*{Acknowledgment}
This work was funded by the National Science Foundation Expanding Frontiers in Research and Innovation Program, grant number 1935278. Special thanks to Ann Grote of the US Fish and Wildlife Service for the lamprey footage included in the online Media. Thanks to Max Austin for helpful conversations regarding hardware as well as matters regarding bioinspiration. Thanks to Hamza Asif and Ash Chase for their help with climbing experiments.


\end{document}